\documentclass[runningheads]{llncs}
\usepackage[T1]{fontenc}
\usepackage{graphicx}
\usepackage{times}
\usepackage{soul}
\usepackage{url}
\usepackage[hidelinks]{hyperref}
\usepackage[utf8]{inputenc}
\usepackage{graphicx}
\usepackage{booktabs}
\usepackage{algorithm}
\usepackage{multirow}
\usepackage[switch]{lineno}
\usepackage{amssymb}
\usepackage{algpseudocode}
\usepackage{amsmath}
\usepackage{cite}
\begin{document}
\title{Neural Architecture Search of Sample Reweighting Networks for Complex Distribution Shift}
\titlerunning{NAS of Sample Reweighting Networks for Complex Distribution Shift}
%\author{First Author\inst{1}\orcidID{0000-1111-2222-3333} \and
%Second Author\inst{2,3}\orcidID{1111-2222-3333-4444} \and
%Third Author\inst{3}\orcidID{2222--3333-4444-5555}}
%\institute{Princeton University, Princeton NJ 08544, USA \and
%Springer Heidelberg, Tiergartenstr. 17, 69121 Heidelberg, Germany

\author{Keisuke Sugawara \and
Kento Uchida \and 
Shinichi Shirakawa
}

\authorrunning{K. Sugawara et al.}

\institute{Yokohama National University,
Kanagawa, Japan\\
\email{sugawara-keisuke-jk@ynu.jp, \{uchida-kento-fz, shirakawa-shinichi-bg\}@ynu.ac.jp}}

\maketitle              % typeset the header of the contribution
\begin{abstract}
    Sample reweighting is a major approach to addressing distribution shifts, such as label noise and class imbalance. Meta-Weight-Net (MW-Net) is a promising sample reweighting network that computes weights based on classification loss. Although MW-Net improves prediction performance under a single type of distribution shift using a simple neural network, its performance degrades when facing both label noise and class imbalance, where it is hard to determine appropriate weights solely from classification loss and using a simple network. In this study, we introduce neural architecture search to MW-Net to mitigate such performance degradation. Using the tree-structured Parzen estimator, we explore the optimal number of hidden layers and nodes and select the most suitable intermediate layer in the classification model to serve as the input for MW-Net. Experimental results on the CIFAR-10 and CIFAR-100 datasets that were modified to include both label noise and class imbalance demonstrate the effectiveness of neural architecture search for MW-Net.
    \keywords{Distribution Shift \and Neural Architecture Search \and Sample Reweighting}
\end{abstract}

\section{Introduction}
Distribution shift refers to the situation in which the distribution of training data differs from that of the operational environment (inference phase). Representative examples of distribution shift include label noise \cite{xiao2015learning} and class imbalance \cite{he2009learning}.
Such shifts often lead to a significant drop in model performance when standard supervised learning methods are directly applied. Consequently, the development of machine learning algorithms that are robust to distribution shifts has become an active area of research.
Sample reweighting is one effective approach to addressing distribution shift. These methods assign weights to individual training samples to reduce the influence of noisy data or overrepresented classes. For example, Focal Loss \cite{lin2020focal} dynamically adjusts sample weights based on the loss values of a classification model. However, due to the variety of weighting functions based on different principles, choosing an appropriate function and tuning its hyperparameters requires expert knowledge, posing a challenge for practical use.

To overcome this issue, recent approaches have proposed to compute the sample weights using machine learning models. 
A representative example is Meta-Weight-Net (MW-Net)~\cite{shu2019meta}. 
MW-Net determines the weight for each sample using the loss value of the classification model.
The update of MW-Net is performed in a meta-learning fashion, which utilizes a small but clean and balanced meta-dataset sampled from the target distribution.
The structure of MW-Net in~\cite{shu2019meta} is a simple fully connected network with one hidden layer of 100 nodes.
Despite its simplicity, MW-Net has demonstrated strong performance in datasets with label noise and class imbalance.

In datasets with label noise, a weighting network that assigns smaller weights to high-loss samples is desirable. Conversely, in class-imbalanced settings, minority class samples often have higher loss due to their learning difficulty, and thus, assigning larger weights to such samples is more effective. Therefore, for a single type of distribution shift, a loss-based weighting network is generally sufficient.
However, in more complex settings where label noise and class imbalance coexist \cite{lu2023label}, both noisy samples (which should be downweighted) and minority class samples (which should be upweighted) tend to exhibit high loss values. In such cases, it becomes difficult to infer appropriate weights based solely on loss.

To address this challenge, we propose the following two modifications to MW-Net:
\begin{enumerate}
    \item We replace the input of MW-Net from the loss value with a combination of the intermediate feature representations from the classifier and the one-hot encoded label.
    \item We jointly optimize both the network architecture of the weighting network (depth and number of nodes) and the feature extraction layer used as input of MW-Net.
\end{enumerate}
The first modification enables the network to assign weights using richer information beyond just the loss, by incorporating semantic features and label context. However, this increase in input dimensionality may render the original MW-Net architecture (a single hidden layer of 100 nodes) insufficient.
Thus, we address this by optimizing the network structure along with the input source using Tree-structured Parzen Estimator (TPE) \cite{bergstra2011algorithms}, which is expected to improve robustness in complex distribution shift scenarios.

We evaluate our method on CIFAR-10 and CIFAR-100~\cite{krizhevsky2009learning} datasets augmented with synthetic label noise and class imbalance. Experimental results show that our method consistently outperforms existing approaches across a wide range of settings.

\section{Related Work}

\subsection{Sample Weighting}

Sample reweighting techniques have been widely explored to mitigate the influence of label noise and class imbalance in supervised learning. 
Although sample reweighting is effective in both label noise and class imbalance, these situations were independently investigated because reasonable weighting strategies are different.
Focal Loss~\cite{lin2020focal} addresses class imbalance by down-weighting easy-to-classify (often majority class) samples, encouraging the model to focus on harder (minority class) examples. Extensions and applications of this idea have been proposed in various domains~\cite{zhong2022mask,NEURIPS2020_f0bda020,mukhoti2020calibrating,tao2023dual}.  
For label noise, self-paced learning~\cite{kumar2010self} and loss-thresholding approaches~\cite{zhang2018generalized} have been proposed to emphasize samples with lower losses, under the assumption that such samples are more likely to be clean.

These approaches typically rely on manually designed weighting functions, requiring domain expertise and careful tuning. To address this issue, MW-Net~\cite{shu2019meta} introduced a meta-learning framework in which a separate weighting network is trained to assign sample weights by minimizing the loss on a small, clean meta-dataset. MW-Net uses a fixed, single-hidden-layer fully connected network and takes loss values as input. Despite its simplicity, it achieves strong performance under label noise and class imbalance.  
To extend this idea, CMW-Net~\cite{shu2023cmw} was proposed to output sample weights at the class-group level, explicitly modeling class-dependent bias in real-world data.
However, MW-Net and CMW-Net were not evaluated under the complex distribution shift composed of label noise and class imbalance.

\subsection{Neural Architecture Search}

Neural Architecture Search (NAS) aims to automate the design of neural network architectures. Early NAS approaches based on reinforcement learning~\cite{zoph2016neural} and evolutionary algorithms~\cite{real2019regularized,Suganuma2017} demonstrated promising results, but required substantial computational resources.  
Several studies have attempted to formalize how network architecture affects robustness to label noise~\cite{labelnoise:nas:2021}, and have proposed improvements to the loss functions of NAS algorithms to mitigate the impact of label noise~\cite{labelnoise:nas:2022}. In addition, enhancements to the loss functions of NAS algorithms have also been proposed to alleviate the effects of class imbalance~\cite{timofeev2021selfsupervisedneuralarchitecturesearch, yao2024efficientnasbasedapproachhandling}.
However, they only considered the architecture of the classification model and did not use the sample weighing network. 
In contrast, we apply NAS to obtain the suitable architecture of the sample weighing network to simultaneously deal with both label noise and class imbalance.
To the best of our knowledge, this is the first study to apply NAS to the sample weighting networks in this meta-learning setting.

\section{Meta-Weight-Net}
\label{sec:mwnet}
Shu et al.~\cite{shu2019meta} proposed a training method for a classification model that is robust to distribution shift by simultaneously training a classification model $f$ with parameters $\mathbf{w}$ and MW-Net $\mathcal{V}$ with parameters $\Theta$ that outputs the weight in $[0,1]$ for each sample. This training method uses two datasets: a training dataset $\mathcal{D}_\mathrm{train}$ consisting of $N$ samples that follow a data distribution different from the operational environment, and a meta-dataset $\mathcal{D}_\mathrm{meta}$ consisting of $M \ll N$ samples that follow the same distribution as the operational environment. 
In this training method, the classification model is updated to minimize the weighted average loss $\mathcal{L}^{\mathrm{train}}$ with the weights computed by the sample weighting network as
\begin{align}
\mathcal{L}^{\mathrm{train}}(\mathbf{w};\Theta) = \frac{1}{N} \sum_{i=1}^N \mathcal{V}(\phi(x_i, y_i; \mathbf{w}) ;\Theta) \cdot L_i^{\mathrm{train}}(\mathbf{w}) \enspace,
\label{classifier}
\end{align}
where $L_i^{\mathrm{train}}(\mathbf{w}) = \ell(y_i, f(x_i; \mathbf{w}))$ is the loss for $i$-th training data, and $\phi(x_i, y_i; \mathbf{w})$ is the input for MW-Net, which is set as $\phi(x_i, y_i; \mathbf{w}) = L_i^{\mathrm{train}}(\mathbf{w})$ in~\cite{shu2019meta}.
On the other hand, the training of MW-Net is designed to minimize the average loss $\mathcal{L}^{\mathrm{meta}}$ for meta-dataset after training of the classification model as
\begin{align}
&\mathcal{L}^{\mathrm{meta}}(\Theta) = \frac{1}{M}\sum_{i=1}^M L_i^{\mathrm{meta}}(\mathbf{w}^*(\Theta)) \label{sample weight} \enspace, \\
&\text{where} \quad \mathbf{w}^*(\Theta) = \underset{\mathbf{w}} {\operatorname{argmin}} \enspace \mathcal{L}^{\mathrm{train}}(\mathbf{w};\Theta) 
\label{eq:optimal:classifier}
\end{align}
and $L_i^{\mathrm{meta}}(\mathbf{w}) = \ell(y_i^{\mathrm{meta}}, f(x_i^{\mathrm{meta}}; \mathbf{w}))$ represents the loss value for the $i$-th meta sample computed with the loss function $\ell$.

To reduce the computational cost of the above-mentioned meta-learning process, each parameter is updated by stochastic gradient descent using mini-batches containing $n$ training samples and $m$ meta samples to minimize each average loss. In addition, instead of using the optimal classifier parameters $\mathbf{w}^\ast(\Theta)$ in Eq.~\eqref{eq:optimal:classifier}, the temporarily updated parameters $\hat{\mathbf{w}}^t(\Theta)$ are computed as 
\begin{align}
\hat{\mathbf{w}}^t(\Theta) &= \mathbf{w}^t - \frac{\eta_\mathbf{w}}{n}\sum_{i =1}^n\mathcal{V}(\phi(x_i, y_i; \mathbf{w}^t); \Theta) \cdot \nabla_\mathbf{w}L_i^{\mathrm{train}}(\mathbf{w})\Big|_{\mathbf{w}^t} \enspace, \label{temporary} 
\end{align}
where $\eta_\mathbf{w}$ denotes the learning rate for the classifier.
Subsequently, each parameter is updated using $\hat{\mathbf{w}}^t(\Theta)$ as 
\begin{align}
\Theta^{t+1} &= \Theta^t - \frac{\eta_\Theta}{m}\sum_{i=1}^m\nabla_{\Theta}L_i^{\mathrm{meta}}(\hat{\mathbf{w}}^t(\Theta)) \Big|_{\Theta^t} \label{update sample weight}
\\
\mathbf{w}^{t+1} &= \mathbf{w}^t - \frac{\eta_\mathbf{w}}{n}\sum_{i=1}^n\mathcal{V}(\phi(x_i, y_i; \mathbf{w}^t); \Theta^{t+1}) \cdot \nabla_\mathbf{w}L_i^{\mathrm{train}}(\mathbf{w})\Big|_{\mathbf{w}^t} \enspace. \label{update classifier}
\end{align}
where $\eta_\Theta$ denotes the learning rate for MW-Net.

\section{Proposed Method}

This study aims to improve performance under complex distribution shifts involving both label noise and class imbalance by optimizing the network structure and input features of MW-Net. In this section, we describe the search space, search algorithm, and evaluation method for architectures used for the neural architecture search of MW-Net.
The algorithm of the proposed method is presented in Algorithm \ref{algorithm MW-Net TPE}.

\subsection{Search Space}
Considering that the MW-Net was originally given by the fully-connected neural network, we optimize its number of layers and number of nodes.
In addition, to determine suitable sample weights under complex distribution shift, we also optimize the input features of MW-Net.
The candidates of the input feature are computed with the output $b_j$ of the selected block of the classification model.
The output $b_j$ is subjected to the global average pooling $\psi_{\mathrm{ave}}$ and concatenated with the one-hot vector $\psi_{\mathrm{onehot}}(y_i)$ representing the label information to form the input vector to MW-Net as
\begin{align}
\phi(x_i, y_i; \mathbf{w}) = ( \psi_{\mathrm{ave}}(b_j(x_i)), \psi_{\mathrm{onehot}}(y_i) ) \enspace.
\end{align}

Table~\ref{tab:propose space} shows the search space used in the experiment in Section~\ref{sec:experiment}. The number of layers is selected from candidates ranging from 1 to 5, while the number of nodes for each layer is chosen from candidates ranging from 64 to 1024 and optimized on a logarithmic scale to improve the efficiency of the search.
The candidate inputs are the last five blocks of the classification model.
In this paper, we use ResNet-32~\cite{he2016deep} as the classification model, which consists of 15 blocks, each composed of convolutional layers, batch normalization, and ReLU activation functions. The candidate inputs are from the 11th to the 15th blocks in ResNet, where the selected output is aggregated as a 64-dimensional feature vector by the global average pooling.

Based on the selected structure parameter, MW-Net is constructed as a fully-connected neural network, in which the ReLU function is used as the activation function for the hidden layers, and the sigmoid function is applied to the output layer.

\begin{algorithm*}[t]
\caption{Architecture Search for MW-Net}\label{algorithm MW-Net TPE}
 \renewcommand{\algorithmicrequire}{\textbf{Input:}}
 \renewcommand{\algorithmicensure}{\textbf{Output:}}
\begin{algorithmic}[1]
    \Require training dataset $\mathcal{D}_{\mathrm{train}}$, meta-dataset $\mathcal{D}_{\mathrm{meta}}$, validation dataset $\mathcal{D}_{\mathrm{val}}$, training batch size $n$, meta batch size $m$, total epochs $E$, number of iterations $T$, number of architecture evaluations $J$, number of architecture evaluations for random search $J_{\mathrm{RS}}$, threshold $\gamma$ in TPE
    \Ensure Best architecture $\mathbf{A}^*$ and its optimal parameters $\mathbf{w}_{\mathbf{A}}^*$
    \For{$j=0, \cdots, J$}
        \If{$j < J_{\mathrm{RS}}$}
            \State Determine architecture $\mathbf{A}$ uniformly randomly on the search space.
        \Else
            \State Divide $\mathbb{D} = \{(\mathbf{A}_k, f_k)\}_{k=0}^{j}$ into upper set $\mathbb{D}^\ell$ and lower set $\mathbb{D}^g$ using threshold $f^\gamma$
            \State Perform kernel density estimation on $\mathbb{D}^\ell$ and $\mathbb{D}^g$
            \State Determine architecture $\mathbf{A}$ that maximizes the density ratio in Eq.~(\ref{tpe EI})
        \EndIf
        \State Initialize classifier parameters $\mathbf{w}$ and sample weighting network parameters $\Theta$
        \For{$e=0, \cdots, E-1$}
            \For{$t=0, \cdots, T-1$}
                \State Sample mini-batch of size $n$ from $\mathcal{D}_{\mathrm{train}}$
                \State Sample mini-batch of size $m$ from $\mathcal{D}_{\mathrm{meta}}$
                \State Update classifier and MW-Net using Eq.~\eqref{temporary}, \eqref{update sample weight}, and \eqref{update classifier}
            \EndFor
            \State Evaluate the accuracy on $\mathcal{D}_{\mathrm{meta}}$
        \EndFor
        \State Evaluate the accuracy $f$ on $\mathcal{D}_{\mathrm{val}}$ using the best model parameter for $\mathcal{D}_{\mathrm{meta}}$ 
        \State $\mathbb{D} \leftarrow \mathbb{D} \cup \{(\mathbf{A}, f)\}$
    \EndFor \\
    \Return Best architecture $\mathbf{A}^*$ and its optimal parameters $\mathbf{w}_{\mathbf{A}}^*$
\end{algorithmic}
\end{algorithm*}

\begin{table}[t]
\centering
\caption{Search Space of the Proposed Method} \label{tab:propose space}
\begin{tabular}{c|c|c}
\hline
& Search Space & Variable Type \\
\hline
Number of Layers
& $\{1,2,3,4,5\}$ & Integer \\
Number of Nodes for each Layer
& $[64, 1024]$ & Integer (log-scale) \\
Input (Block)
& $\{11,12,13,14,15\}$ & Integer \\
\hline
\end{tabular}
\end{table}

\subsection{Search Algorithm}
In this study, the input selection and neural architecture search for MW-Net are performed using the Tree-structured Parzen Estimator (TPE) \cite{bergstra2011algorithms}. 
TPE is a method based on Bayesian optimization, which uses a set of evaluated solutions and their corresponding evaluation values, denoted as $\mathbb{D} = {(\mathbf{A}_k, f_k)}_{k=1}^{K}$, to determine the next solution $\mathbf{A}$ to be evaluated. 

For the maximization of the evaluation value, TPE calculates the $(1-\gamma)$-quantile $f^\gamma$ of $\mathbb{D}$ and divides the set $\mathbb{D}$ into two subsets: the upper set $\mathbb{D}^\ell$ containing the structures with evaluation values greater than or equal to $f^\gamma$ and the lower set $\mathbb{D}^g$ containing the structures with evaluation values less than $f^\gamma$. 
Then, TPE assumes that the probability density function is given by the kernel density estimation $p(\mathbf{A}|\mathbb{D}^\ell)$ and $p(\mathbf{A}|\mathbb{D}^g)$ for each set, i.e.,
\begin{align}
p(\mathbf{A}|f, \mathbb{D}) = \begin{cases}
p(\mathbf{A}|\mathbb{D}^\ell) & (f \ge f^\gamma) \\
p(\mathbf{A}|\mathbb{D}^g) & (f < f^\gamma)
   \end{cases}
   \enspace. \label{eq:tpe:density}
\end{align}

In Bayesian optimization, the maximizer of the acquisition function is selected as the next candidate point. Expected Improvement (EI) and Probability of Improvement (PI) are widely used acquisition functions, which are defined as
\begin{align}
\mathrm{EI}_{f^\gamma}[\mathbf{A} \mid \mathbb{D}] = \int^{\infty}_{f^\gamma}(f-f^\gamma)p(f|\mathbf{A}, \mathbb{D})\mathrm{d}f \\
\Pr(f \geq f^\gamma \mid \mathbf{A}, \mathbb{D}) = \int^{\infty}_{f^\gamma} p(f|\mathbf{A}, \mathbb{D})\mathrm{d}f \enspace.
\end{align}
Under the probability density function in~\eqref{eq:tpe:density}, both the maximization of EI and PI can be reformulated into the maximization of the ratio between the density functions on the upper and lower sets.
More precisely, the rankings of the structure parameters on these acquisition functions are consistent as follows:
\begin{align}
    \mathrm{EI}_{f^\gamma}[\mathbf{A}|\mathbb{D}] \stackrel{\mathrm{rank}}{\simeq} \Pr(f \geq f^\gamma \mid \mathbf{A}, \mathbb{D}) \stackrel{\mathrm{rank}}{\simeq} \frac{p(\mathbf{A}|\mathcal{\mathbb{D}^\ell})}{p(\mathbf{A}|\mathbb{D}^g)} \enspace. \label{tpe EI}
\end{align}
Based on this, the structure parameters that maximize the density ratio are selected as the next structure to evaluate. 

We implemented TPE using \texttt{Optuna 4.0.0}~\cite{optuna} with default hyperparameter settings. Before performing TPE, the algorithm performs a random search for $J_\mathrm{RS} = 10$ iterations.

\subsection{Evaluation of Architecture}

After the generation of the structure parameters $\mathbf{A}$, we train both the classifier and MW-Net with the training process explained in Section~\ref{sec:mwnet}.
At the end of each epoch, we evaluated the accuracy of the classifier with the meta-dataset $\mathcal{D}_\mathrm{meta}$.
Then, after the training, we evaluated the classifier 
 with the best parameters for the meta-dataset using the validation dataset $\mathcal{D}_\text{val}$, which follows the same distribution as the meta and test datasets. 
Then, the accuracy on the validation dataset is used as the evaluation value of the structure parameters in TPE.

We note that, in practice, the validation dataset can be prepared by divining the meta-dataset into two folds.
Therefore, less metadata is used in the training of MW-Net when applying architecture search.

\section{Experiments}
\label{sec:experiment}
In this section, we evaluate the effectiveness of the proposed method through experiments on benchmark datasets.
Section \ref{Dataset Construction} describes the construction of datasets with imbalanced and noisy labels.
Section \ref{Experimental Settings} outlines the experimental settings, including hyperparameters and model configurations.
Sections \ref{Results on CIFAR-10} and \ref{Results on CIFAR-100} present the experimental results on CIFAR-10 and CIFAR-100 \cite{krizhevsky2009learning}, respectively, under various noise and imbalance conditions.

\subsection{Dataset Construction} \label{Dataset Construction}
This section describes the procedure for constructing a dataset with label noise and class imbalance.
First, class imbalance is introduced by adjusting the number of samples per class according to the following equation.
\begin{equation*}
n_i = \frac{n}{\beta^{\frac{i}{c-1}}}
\end{equation*}
Here, $n$ denotes the original number of samples per class, $c$ is the total number of classes, and $n_i$ represents the number of samples in the $i$-th class.
The parameter $\beta$ denotes the Imbalance Factor (IF), which indicates how many times larger the number of samples in the largest class is compared to the smallest class.
Generally, a higher IF corresponds to a more challenging classification task.
In this study, we consider two levels of imbalance: $\beta = 20$ and $\beta = 100$.

CIFAR-10 and CIFAR-100 datasets consist of 50,000 training images and 10,000 test images, with 10 and 100 classes, respectively. The number of samples is evenly distributed across all classes. When class imbalance is introduced into the training data, the number of samples is reduced to 16,683 and 12,156 for CIFAR-10 under IF20 and IF100, respectively, and to 15,594 and 10,626 for CIFAR-100 under IF20 and IF100, respectively.

After introducing class imbalance, label noise is added to the dataset.
In this study, we consider two types of noise: {\it uniform noise} and {\it flip noise}.
In the case of uniform noise, data samples from a given class are uniformly randomly reassigned to any of the classes in the dataset. This type of noise simulates random labeling errors that may occur during the data collection process.
On the other hand, flip noise changes the label of a sample to a pre-specified different class with a certain probability.
In this study, we set the noise rate to 40\%.

\begin{figure}[t]
  \centering
  \includegraphics[width=0.5\linewidth]{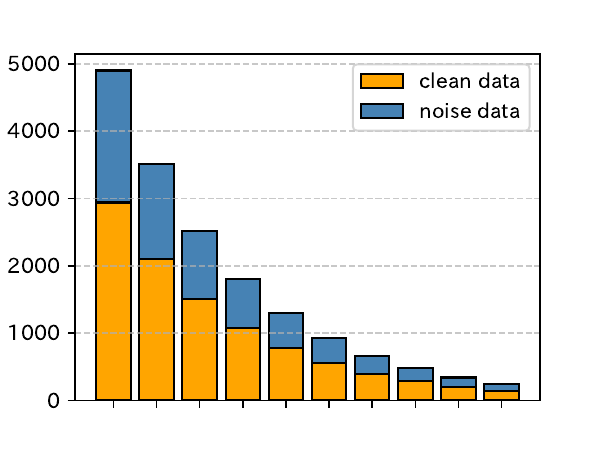}
  \caption{Class-wise distribution of the dataset containing class imbalance and label noise. ``Noise data" refers to samples assigned with incorrect class labels. The imbalance factor (IF), $\beta = 20$, and label noise rate of 40\% are used.}
  \label{imbalance_and_noise}
\end{figure}

Figure~\ref{imbalance_and_noise} shows the class-wise distribution of the dataset after applying the aforementioned procedures.
In this figure, the noise data represent samples that have been assigned incorrect class labels.

\subsection{Experimental Settings} \label{Experimental Settings}
This section describes the experimental settings.
\begin{table*}[t]
    \centering
    \caption{Training Settings and Optimization Algorithms}
    \label{tab:settings}
    \begin{tabular}{ll}
        \hline
        \multicolumn{2}{l}{\textbf{Classification Model}} \\ \hline
        Model & ResNet-32~\cite{he2016deep} \\ 
        Learning Algorithm & SGD \\ 
        Batch Size & 100 \\ 
        Initial Learning Rate & 0.1 \\ 
        Momentum Coefficient & 0.9 \\ 
        Weight Decay & $5\times10^{-4}$ \\
        Total Epochs & 100 \\ 
        Learning Rate Decay Epochs & 80, 90 \\ \hline \hline \addlinespace
        \multicolumn{2}{l}{\textbf{Sample Weighting Network}} \\ \hline
        Input Layer Dimension & Feature dimension + one-hot label representation \\ 
        Learning Algorithm & Adam $(\beta_1 = 0.9, \beta_2 = 0.999)$ \\
        Learning Rate & $1\times10^{-5}$ \\
        Weight Decay & $1\times10^{-4}$ \\
        Activation Function (Hidden Layers) & ReLU \\ 
        Activation Function (Output Layer) & Sigmoid \\ \hline \hline \addlinespace
        \multicolumn{2}{l}{\textbf{TPE}} \\ \hline
        Threshold Score ($\gamma$) & 10\% \\
        Number of Evaluations & 100 \\
        Evaluation Metric & Top-1 accuracy on validation data \\ \hline
    \end{tabular}
\end{table*}
The training configurations and optimization settings for the models used in the experiments are summarized in Table~\ref{tab:settings}.

As baseline methods, we evaluated two architectures: a conventional architecture with a single layer of 100 nodes and the largest architecture in the search space consisting of 5 layers with 1024 nodes. In both cases, the input was constructed by concatenating the output from the 15th block of the classification model with the one-hot encoding of the label.
Since these baseline methods do not require the structure validation dataset $\mathcal{D}_\text{val}$, we merged that data into the meta-dataset $\mathcal{D}_\mathrm{meta}$.
We also evaluated a setting where the input was fixed as the output of the 15th block of the classifier, and only the number of layers and nodes was searched.

Experiments were conducted under two conditions: one where both the meta-dataset and the validation dataset consisted of 500 samples, and another where they consisted of 100 samples each. For the baseline methods without architecture search, the meta-dataset size was set to 1,000 or 200 samples, respectively.

In the experiments using CIFAR-100, we only evaluated the setting with 500 samples each for the meta-dataset and the structure validation dataset, while the baseline method used 1,000 meta samples.
For both CIFAR-10 and CIFAR-100, we conducted three independent runs per setting.

Notably, the meta-dataset, structure validation dataset, and test dataset are class-balanced and free from label noise. Therefore, using Top-1 accuracy as the evaluation metric does not pose a significant problem.

\begin{table*}[t]
\centering
\caption{Test accuracy on CIFAR-10 with 1,000 meta samples (average $\pm$ standard deviation)}
\label{result cifar10 meta1000}
\renewcommand{\arraystretch}{1.2}
\setlength{\tabcolsep}{2pt}
\begin{tabular}{r|cccc}
\hline
Setting & Baseline & Largest Arch. & NAS (Arch. Only) & NAS (Arch. and Input) \\ \hline
\multirow{1}{*}[0pt]{\shortstack{IF20/Flip40}}
 & 73.75{\footnotesize$\pm$0.48} & 75.43{\footnotesize$\pm$1.73} & 76.72{\footnotesize$\pm$1.37} & \textbf{77.42}{\footnotesize$\pm$1.62} \\ 
\multirow{1}{*}[0pt]{\shortstack{IF100/Flip40}}
 & 59.88{\footnotesize$\pm$3.06} & 65.29{\footnotesize$\pm$2.77} & 66.72{\footnotesize$\pm$0.40} & \textbf{67.30}{\footnotesize$\pm$1.31} \\
\multirow{1}{*}[0pt]{\shortstack{IF20/Uniform40}}
 & 74.36{\footnotesize$\pm$1.31} & 73.20{\footnotesize$\pm$0.63} & 75.07{\footnotesize$\pm$0.60} & \textbf{75.89}{\footnotesize$\pm$0.83} \\
\multirow{1}{*}[0pt]{\shortstack{IF100/Uniform40}}
 & 55.22{\footnotesize$\pm$2.97} & 58.03{\footnotesize$\pm$4.93} & \textbf{61.50}{\footnotesize$\pm$3.88} & 61.41{\footnotesize$\pm$3.81} \\ \hline
\end{tabular}
\end{table*}

\begin{table*}[t]
\centering
\caption{Test accuracy on CIFAR-10 with 200 meta samples (average $\pm$ standard deviation)}
\label{result cifar10 meta200}
\renewcommand{\arraystretch}{1.2}
\setlength{\tabcolsep}{2pt}
\begin{tabular}{r|cccc}
\hline
Setting & Baseline & Largest Arch. & NAS (Arch. Only) & NAS (Arch. and Input) \\ \hline
\multirow{1}{*}[0pt]{\shortstack{IF20/Flip40}}
& 71.05{\footnotesize$\pm$1.78} & 73.67{\footnotesize$\pm$1.39} & 75.95{\footnotesize$\pm$1.36} & \textbf{76.34}{\footnotesize$\pm$1.73} \\ 
\multirow{1}{*}[0pt]{\shortstack{IF100/Flip40}}
& 59.65{\footnotesize$\pm$1.28} & 64.19{\footnotesize$\pm$2.73} & 65.47{\footnotesize$\pm$3.02} & \textbf{65.54}{\footnotesize$\pm$2.10} \\
\multirow{1}{*}[0pt]{\shortstack{IF20/Uniform40}}
& 73.80{\footnotesize$\pm$1.70} & 73.29{\footnotesize$\pm$0.61} & 73.54{\footnotesize$\pm$0.24} & \textbf{74.05}{\footnotesize$\pm$0.27} \\
\multirow{1}{*}[0pt]{\shortstack{IF100/Uniform40}}
& 56.79{\footnotesize$\pm$5.11} & 54.99{\footnotesize$\pm$4.06} & 60.29{\footnotesize$\pm$4.35} & \textbf{61.15}{\footnotesize$\pm$3.22} \\
\hline
\end{tabular}
\end{table*}
\subsection{Results on CIFAR-10} \label{Results on CIFAR-10}
\paragraph{\textbf{Quantitative Comparison of Architectures}}\mbox{}\\
\indent Tables~\ref{result cifar10 meta1000} and \ref{result cifar10 meta200} show the average and standard deviation of Top-1 accuracy on the test dataset when the number of meta samples is set to 1,000 and 200, respectively.
The proposed methods correspond to the entries labeled ``NAS (Arch. Only)" and ``NAS (Arch. and Input)" in the tables.
Across all problem settings and regardless of the size of the meta-dataset, the architecture search consistently yielded the best performance.
Moreover, a general trend can be observed: the maximum architecture outperformed the baseline, and the architecture search outperformed both.

\begin{figure}[t]
    \begin{center}
    \includegraphics[width=0.99\textwidth]{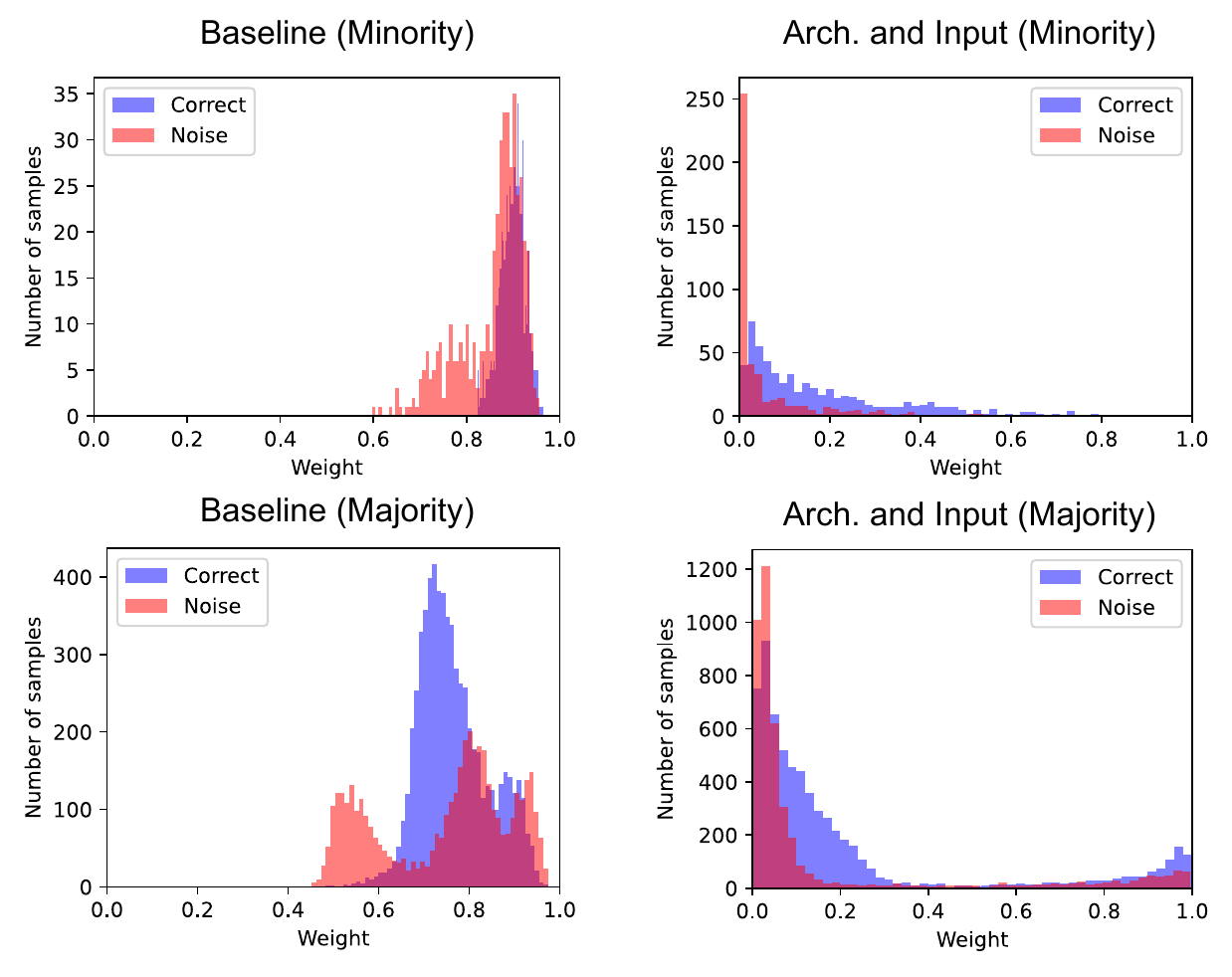}
    \caption{
    The distributions of the sample weights on CIFAR-10 with IF20 and 40\% Flip noise for containing 1,000 meta samples. 
    Top: sample weights for minority classes (the three classes with the fewest samples).  
    Bottom: sample weights for majority classes (the three classes with the most samples).  
    The left panels correspond to the baseline architecture, and the right panels show the results when architecture search (with the input selection) is applied.
    }
    \label{fig: weight_domain}
    \end{center}
\end{figure}

\paragraph{\textbf{Examination of Sample Weight Distributions}}\mbox{}\\
\indent Figure~\ref{fig: weight_domain} shows the histograms of sample weights assigned to training samples belonging to minority and majority classes under both the baseline architecture and the architecture search with the input selection.
The minority classes are defined as the three classes with the fewest number of samples based on ground-truth labels, while the majority classes are defined as the three classes with the most samples.
For each case, histograms are drawn separately for noisy samples (in red) and clean samples (in blue).

For the minority classes, under the baseline architecture, high sample weights (greater than 0.6) are assigned regardless of the presence of label noise, indicating that noisy data cannot be effectively distinguished.
In contrast, when architecture search is applied, noisy samples are assigned weights close to zero in most cases, while clean samples receive significantly higher weights, demonstrating that the model is able to differentiate between the two.
This appropriate assignment of sample weights to the harder-to-learn minority classes likely contributes to the improvement in classification performance.

For the majority classes, a similar trend is observed under the baseline architecture, where noisy samples are again assigned high weights.
With the architecture search, noisy samples tend to receive lower weights, suggesting some level of distinction.
However, it is also observed that many samples---regardless of noise---are still assigned weights close to 1.0.
Since majority classes are typically overrepresented, assigning lower weights to them overall may further enhance performance.

\begin{figure}[t]
    \begin{center}
    \includegraphics[width=0.99\textwidth]{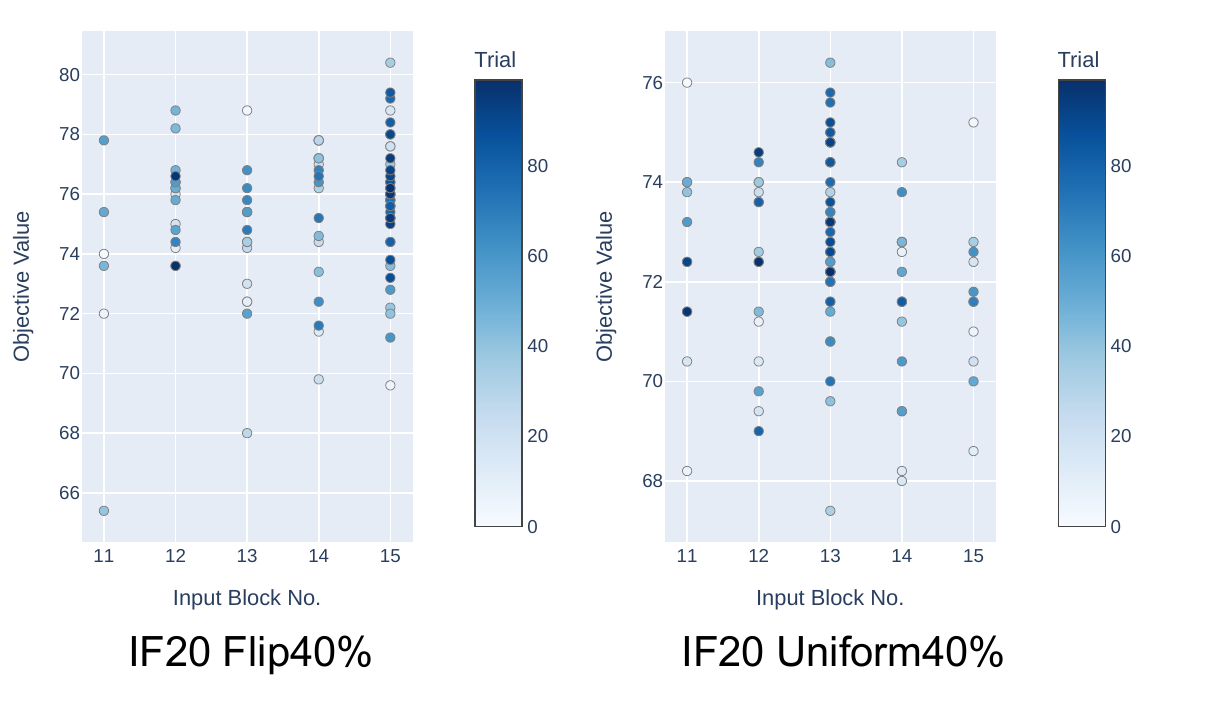}
    \caption{Search history of feature extraction positions from the classifier in architecture and input feature selection.  
    The left panel shows the results under IF $\beta=20$ and 40\% flip noise, while the right panel shows the results under IF $\beta=20$ and 40\% uniform noise. The vertical axis represents the Top-1 accuracy for the variation dataset.}
    \label{fig: get_features}
    \end{center}
\end{figure}

\paragraph{\textbf{Evaluation of the Effect of Input Feature Selection}}\mbox{}\\
\indent Figure~\ref{fig: get_features} visualizes the search history of feature extraction positions from the classifier during the optimization of both architecture and input feature selection.
This represents the MW-Net configurations evaluated in the best trial under the condition of 1,000 meta samples.
From the left panel of the figure, it can be seen that in the case of flip noise, $\mathtt{get\_feature}=15$---corresponding to the final block (15th block) of the classifier---consistently received the highest evaluation scores and was frequently selected in the later stages of the search.
For all experimental settings with the flip noise, the input features were often extracted from the 14th or 15th layer, suggesting a tendency under the flip noise to prefer features from the final block or its vicinity. 
Tables~\ref{result cifar10 meta1000} and~\ref{result cifar10 meta200} show that slightly higher test accuracy is obtained when applying the input selection. Because the 15th layer is used when not applying the input selection, the small difference in the extracted position makes such a performance improvement under the flip noise.

In contrast, the right panel of the figure shows that, under the uniform noise, there were several solutions in which features were selected from earlier layers, such as the 11th, 12th, and 13th blocks.
Under the uniform noise, there appears to be a tendency to prefer features from the earlier stages of the classifier. Moreover, slight performance improvement was observed in Tables~\ref{result cifar10 meta1000} and~\ref{result cifar10 meta200}, compared to the architecture search without input feature selection, which may be attributed to this difference in the feature extraction position.

\begin{figure}[t]
    \begin{center}
    \includegraphics[width=0.99\textwidth]{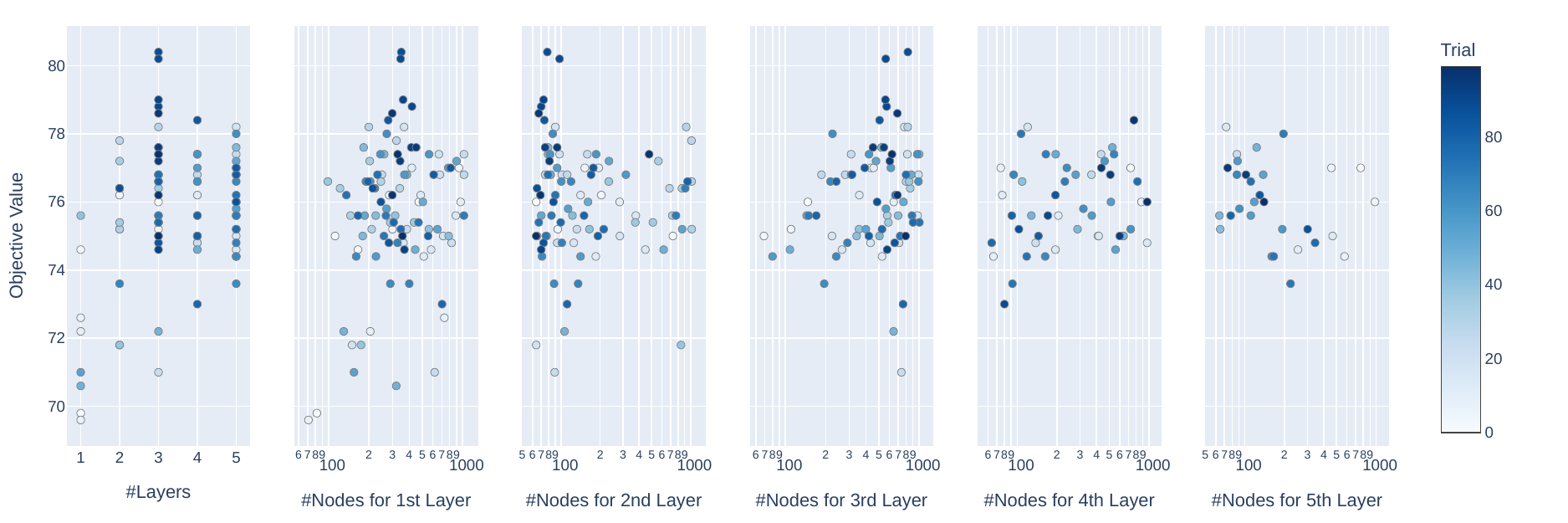}
    \caption{
    Neural Architecture search results under the IF20, Flip 40\% setting with 1,000 meta samples
    }
    \label{fig: architecture_flip}
    \end{center}
\end{figure}

\begin{figure}[t]
    \begin{center}
    \includegraphics[width=0.99\textwidth]{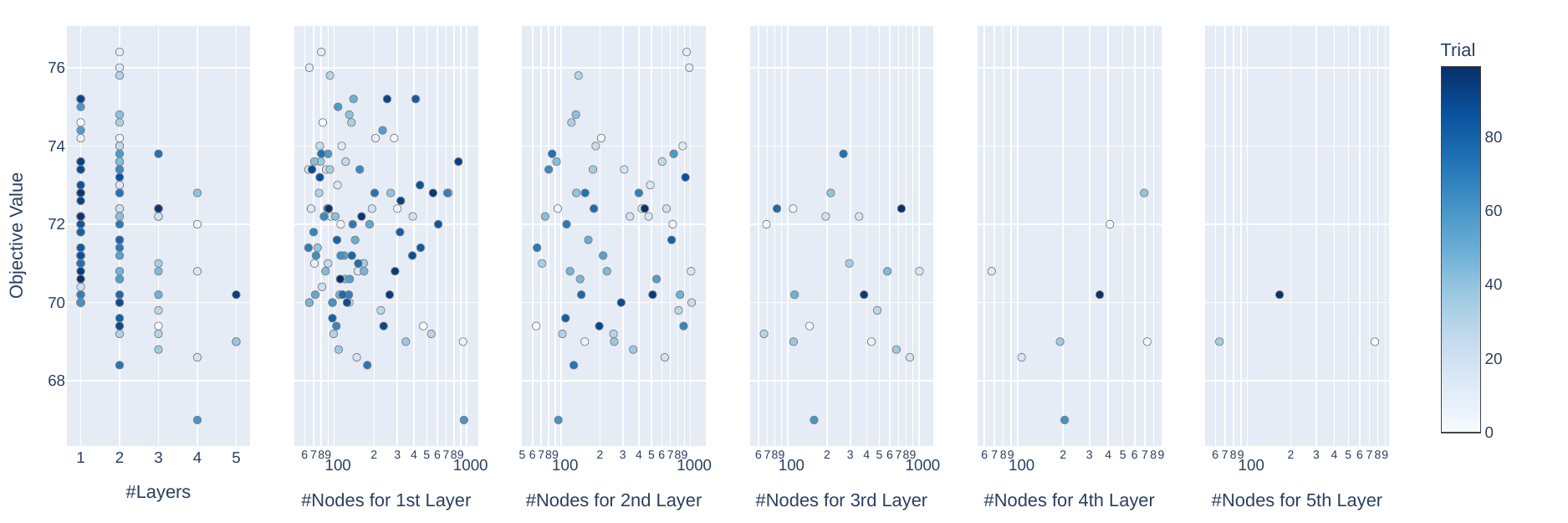}
    \caption{
    Neural Architecture search results under the IF20, Uniform 40\% setting with 1,000 meta samples
    }
    \label{fig: architecture_unif}
    \end{center}
\end{figure}

\paragraph{\textbf{Analysis of the Obtained Architectures}}\mbox{}\\
\indent Figures~\ref{fig: architecture_flip} and~\ref{fig: architecture_unif} present the architecture search results, including the search trajectories, under the IF20/Flip40 and IF20/Uniform40 noise settings, respectively.
From Figure~\ref{fig: architecture_flip}, it can be observed that architectures with three to more than five hidden layers were frequently explored during the later stages of the search in the flip noise setting.
The best-performing architecture had a three-layer structure, with the number of nodes transitioning as 349--78--824.
These results suggest that deeper architectures with three or more layers tend to be favored under the flip noise, which likely explains why they outperform the manually designed baseline structure.

In contrast, Figure~\ref{fig: architecture_unif} shows that, under the uniform noise, architectures with a single hidden layer were predominantly explored during the latter stages.
The best-performing architecture had two layers, where the first layer contained fewer than 100 nodes, while the second had close to 1,000 nodes.
These findings indicate that, in the uniform noise setting, the number of layers tends to remain similar to the baseline, while increasing the number of nodes plays a more critical role.
This architectural tendency may explain why the performance under the uniform noise is closer to that of the baseline compared to the flip noise setting.

We finally discuss the differences between the two noise types and the resulting architectures.
Since the flip noise has fixed label transitions, it likely requires deeper architectures with three or more layers to capture such relationships. 
In contrast, the uniform noise has random label transitions, so architectures with only one or two layers appear sufficient. These interpretations remain at the level of experimental observations, and a rigorous theoretical analysis is left for future work.

\subsection{Results on CIFAR-100} \label{Results on CIFAR-100}
\begin{table*}[t]
\centering
\caption{Test accuracy on CIFAR-100 with 1,000 meta samples (average $\pm$ standard deviation)}
\label{result cifar100 meta1000}
\renewcommand{\arraystretch}{1.2}
\setlength{\tabcolsep}{2pt}
\begin{tabular}{r|cccc}
\hline
Setting & Baseline & Largest Arch. & NAS (Arch. Only) & NAS (Arch. and Input) \\ \hline
IF20/Flip40 & 33.58$\pm$0.20 & 34.18$\pm$0.77 & 34.08$\pm$0.27 & \textbf{34.41}$\pm$0.80 \\
IF100/Flip40 & 25.62$\pm$0.07 & 26.01$\pm$0.31 & \textbf{26.30}$\pm$0.36 & 26.29$\pm$0.57 \\
IF20/Uniform40 & 36.91$\pm$0.88 & 36.94$\pm$0.37 & \textbf{37.36}$\pm$0.45 & 36.78$\pm$1.23 \\
IF100/Uniform40 & 27.44$\pm$0.73 & 26.80$\pm$0.74 & \textbf{27.98}$\pm$1.29 & 27.78$\pm$1.16 \\
\hline
\end{tabular}
\end{table*}

The experimental results on the CIFAR-100 dataset are shown in Table~\ref{result cifar100 meta1000}.
In many settings, both the architecture search with and without the input feature selection outperform the baseline structure.
However, the overall performance gap with the largest architecture is small and, in some combinations of settings and methods, the largest architecture even achieves better accuracy.
These findings suggest that, due to the difficulty of the task in CIFAR-100, the effect of architectural differences on performance is relatively limited.

One possible reason for this phenomenon is the extremely small number of training samples per minority class. In CIFAR-100, the most minority class contains only 25 samples under IF20 and just 5 samples under IF100.
Moreover, since 40\% label noise is added, the number of clean (correctly labeled) samples is further reduced to approximately 15 and 3, respectively.
Under such conditions, where only a handful of clean samples are available, it becomes highly challenging to distinguish between noisy and clean data.
As a result, the difference in performance across methods tends to diminish.

\section{Conclusion}
In this study, we aimed to improve performance on datasets that exhibit two types of distribution shift---label noise and class imbalance---by evaluating the effect of architecture search and input feature selection for MW-Net in order to explore architectures capable of performing more sophisticated sample weighting.
Specifically, we optimized the depth and width of MW-Net, as well as the selection of the intermediate feature layer from the classifier used as input, using the Tree-structured Parzen Estimator (TPE).
Experimental results on CIFAR-10 and CIFAR-100 datasets augmented with synthetic label noise and class imbalance demonstrated that architecture search for MW-Net enables the model to acquire sample weights that effectively suppress performance degradation in the classifier.

Future directions include jointly optimizing the architectures of both MW-Net for sample weighting and the classifier to achieve further performance gains.
Expanding the search space of MW-Net to include convolutional and pooling layers also represents a promising direction.
Furthermore, since the primary purpose of this study was to examine the effect of applying NAS to MW-Net, each candidate architecture was evaluated by training from scratch, without using sophisticated techniques to reduce computational cost.
However, in practice, large-scale datasets must be taken into account, and the current approach is computationally prohibitive.
Therefore, applying state-of-the-art NAS methods or One-Shot NAS methods~\cite{Liu2019darts,Akimoto2019ICML} to reduce the computational overhead of architecture search will be a key direction for future work.

\subsubsection*{Acknowledgements}
This work was partially supported by JSPS KAKENHI Grant Numbers JP23H00491 and JP23H03466, and JST PRESTO Grant Number JPMJPR2133.

\bibliographystyle{splncs04}
\bibliography{mybibliography}
\end{document}